\documentclass[lettersize,journal]{IEEEtran}
\usepackage{cite}
\usepackage{xcolor}
\usepackage{amsmath,amssymb,amsfonts}
\usepackage{algorithmic}
\usepackage{graphicx}
\usepackage{textcomp}
\def\BibTeX{{\rm B\kern-.05em{\sc i\kern-.025em b}\kern-.08em
    T\kern-.1667em\lower.7ex\hbox{E}\kern-.125emX}}

\usepackage{xtab}
\usepackage{graphicx}
\usepackage{array} 
\usepackage{multirow} 
\usepackage[table,xcdraw]{xcolor} 
\usepackage{placeins} 
\usepackage{booktabs}    
\usepackage{ragged2e}
\usepackage{makecell}    
\usepackage{array}   
\usepackage[protrusion=true,expansion=true]{microtype}
\pdfoutput=1
\usepackage{tabularx}
\usepackage{url}
\usepackage[breaklinks=true]{hyperref}

\usepackage{placeins}  
\usepackage{float}     

\definecolor{Gray}{gray}{0.9}
\definecolor{darkyellow}{RGB}{159, 174, 191} 
\definecolor{salmon}{RGB}{208, 222, 196}
\definecolor{darksalmon}{RGB}{171, 196, 150}
\definecolor{darkgreen}{RGB}{180, 230, 198}
\definecolor{lightgreen}{RGB}{209, 233, 201}

\definecolor{lightred}{RGB}{255,211,222}
\definecolor{darkred}{RGB}{252,150,167}
 \definecolor{LightBlue}{rgb}{0.8,0.89,1} 
\definecolor{DarkBlue}{rgb}{0.57,0.71,0.82} 
\definecolor{MagLight}{rgb}{1, 0.89, 0.8}
\usepackage{enumitem}

\usepackage[most]{tcolorbox}

\usepackage{xparse} 

\usepackage{etoolbox} 

\usepackage{nicematrix}

\makeatletter
\tcbset{
  rgb/.code args={#1,#2,#3}{%
    \definecolor{customrgb}{RGB}{#1,#2,#3}%
    \tcbset{colback=customrgb}%
  }
}
\makeatother

\newtcolorbox[auto counter]{Summary}[1][]{
  title={\bfseries Summary},
  enhanced,
  drop shadow={black!50!white},
  coltitle=black,
  top=0.3in,
  attach boxed title to top left={
    xshift=1.5em,
    yshift=-\tcboxedtitleheight/2
  },
  boxed title style={
    size=small,
    colback=pink
  },
  #1
}

  \makeatletter
\newcommand\notsotiny{\@setfontsize\notsotiny\@vipt\@viipt}
\makeatother
\begin{document}

\title{Federated Foundation Models over Vehicular Networks}
\author{
Kasra Borazjani,~\IEEEmembership{Student Member,~IEEE,} Fardis Nadimi,~\IEEEmembership{Student Member,~IEEE,} Payam Abdisarabshali,
~\IEEEmembership{Student~Member,~IEEE,} Owen Palinski,~\IEEEmembership{Student Member,~IEEE,} 
Allan Salihovic,  
Dinh~Nguyen,~\IEEEmembership{Senior Member,~IEEE,}
Minghui Liwang,~\IEEEmembership{Senior Member,~IEEE,} 
and
Seyyedali Hosseinalipour,~\IEEEmembership{Senior Member,~IEEE}
}



\maketitle

\begin{abstract}
 This paper presents a forward-looking vision for integrating the emerging \textit{\underline{m}ulti-\underline{m}odal \underline{m}ulti-\underline{t}ask \underline{fed}erated \underline{f}oundation \underline{m}odels (M3T FedFMs)} into vehicular networks, with the goal of unifying the expressive power of multi-modal multi-task foundation models (M3T FMs) with the privacy-preserving and distributed learning capabilities of federated learning (FL). Given the largely underexplored nature of this research direction, we first introduce the fundamental training/fine-tuning principles of M3T FedFMs. We then discuss a range of their representative use cases in vehicular networks, illustrating the significant potential of M3T FedFMs to enable next-generation vehicular intelligence. Afterwards, 
we identify key constraints inherent to vehicular environments that challenge the practical deployment of M3T FedFMs, and articulate a set of forward-looking research directions to address these challenges. Furthermore, through a case study conducted on a real-world vehicular dataset (i.e., Waymo Open Dataset), we demonstrate the promise of M3T FedFMs for vehicular networks and release our implementation to facilitate reproducibility and stimulate research in this emerging area (repository: \href{https://github.com/KasraBorazjani/vehicular-fedfm}{https://github.com/KasraBorazjani/vehicular-fedfm}). 
\end{abstract}

\maketitle

\section{Introduction}

The field of machine learning (ML) has recently witnessed a major transformation with the advent of \textit{large language models (LLMs)}, which have demonstrated remarkable capabilities in understanding, generation, and reasoning of natural language across a wide range of tasks (e.g., text summarization, translation, and semantic analysis). While LLMs (e.g., GPT 2, and Gemini 1) may be viewed as the most prominent recent breakthrough, ML is simultaneously undergoing a broader and arguably more structural shift driven by the emergence of \textit{\underline{m}ulti-\underline{m}odal \underline{m}ulti-\underline{t}ask \underline{f}oundation \underline{m}odels (M3T FMs)}~\cite{rothman2024transformers}. In particular, M3T FMs (e.g., GPT 5, and Gemini 3) extend the text-centric capabilities of LLMs to jointly process, fuse, and reason over heterogeneous data modalities (e.g., vision, audio, and text), while supporting the concurrent learning of multiple tasks (e.g., image generation, classification, segmentation, and captioning)  within a unified model architecture.

Concurrent with these major innovations in ML, \textit{vehicular networks} are undergoing a paradigm shift driven by the deployment of edge-native ML applications. In particular, vehicles are increasingly envisioned as \textit{mobile edge nodes} equipped with multi-modal sensors, including light detection and ranging (LiDAR), millimeter-wave radar, wide-angle cameras, and global navigation satellite systems (GNSS), and expected to execute multiple downstream tasks, such as 3D object detection, trajectory prediction, and collaborative planning/navigation. \textit{This evolution presents a unique opportunity to integrate M3T FMs within the vehicular networking ecosystem.}
Despite its strong promise, realizing such an integration in practice remains non-trivial: a primary challenge stems from the predominantly \textit{centralized} training/fine-tuning of M3T FMs, which is fundamentally misaligned with the data acquisition realities of vehicular networks. In particular, data in vehicular networks is often geo-distributed across vehicles, roadside units (RSUs), and edge/cloud infrastructure, and centralizing these data for training/fine-tuning M3T FMs may raise privacy concerns (by exposing sensitive sensing, location, or behavioral information to untrusted parties) and impose substantial communication overhead due to the transmission of raw multi-modal data.

Such challenges motivate a shift toward the use of \textit{distributed learning} principles, with the most prominent example being \textit{federated learning (FL)}~\cite{mcmahan2017communication}, for the training/fine-tuning of M3T FMs over vehicular networks. This shift has recently given rise to a new and rapidly emerging research direction in ML, called \textit{\underline{m}ulti-\underline{m}odal \underline{m}ulti-\underline{t}ask \underline{fed}erated \underline{f}oundation \underline{m}odels (M3T FedFMs)}~\cite{ren2025advances}, which enables collaborative training/fine-tuning of M3T FMs across decentralized data sources while preserving data locality.
 However, despite attracting attention in the ML community, 
  \textit{the application of M3T FedFMs in vehicular networks remains largely unexplored}.
  

Motivated by this gap, in this work, we present a vision for integrating M3T FedFMs into vehicular networks, with our contributions summarized as follows: 
(i) We establish a unified architectural perspective
for M3T FedFMs in vehicular networks, elucidating
how pre-trained FMs, modality-specific encoders,
shared representation-learning components,
task-specific heads, and fine-tuning mechanisms can
be integrated to enable distributed model adaptation
across vehicular edge nodes. We further illustrate
the potential of this paradigm through representative
use cases across vehicular ecosystems.
(ii) We identify key challenges arising from the
interplay between M3T FedFMs and the distinctive
characteristics of vehicular networks, including
unpredictable vehicle availability and handoffs,
modality-task heterogeneity, time-varying data,
and privacy risks associated with shared model
updates. We further outline corresponding
research directions to guide the development
of future M3T FedFM frameworks tailored to
vehicular environments.
(iii) We develop and evaluate FedAdapt, a
task-onboarding-oriented M3T FedFM architecture
that extends a pre-trained CLIP model with
multi-modal and multi-task adaptation modules.
Through a case study using the real-world Waymo
Open Dataset, we demonstrate the benefits of
our architectural design and publicly release
the corresponding implementation to facilitate
reproducibility and future research.

\section{Related Work and Background}

\subsection{FL in Vehicular Networks}
Conventional FL proceeds iteratively through three main steps until model convergence: (i) FL devices (e.g., vehicles) independently train local models using their private data; (ii) they periodically transmit model updates (e.g., parameters or gradients) to a coordinating server; and (iii) the server aggregates the received updates (e.g., via weighted averaging) to construct a global model, which is then broadcast back to the clients to synchronize their local models and initiate the next model training round.
By allowing clients to collaboratively train a model without centralizing their raw training data, FL has attracted considerable attention in vehicular networks.
Representative applications include driver monitoring, personalized modeling of individual driving behaviors, advanced driver-assistance systems (ADAS) and steering control, cooperative perception, and traffic prediction/management~\cite{du2020federated}.

\subsection{M3T FMs in Vehicular Systems}
Exploring the applications of various M3T FMs in vehicular networks is a recent research topic. For example, \textit{EMMA (End-to-End Multi-modal Model for Autonomous Driving)} is an M3T FM that integrates camera inputs and navigation instructions to produce outputs such as planner trajectories, perception objects, and road graph elements~\cite{hwang2024emma}. 
Also, 
DeepInteraction++ is an M3T FM  for autonomous driving~\cite{yang2024deepinteraction} with exceptional performance in 3D object detection and end-to-end driving tasks. 
Collectively, these studies demonstrate the potential of integrating M3T FMs into vehicular networks; however, they rely on centralized training/fine-tuning of M3T FMs, where data is first collected at a centralized location (e.g., a cloud server) prior to model training. Such an assumption may not naturally extend to realistic vehicular networks, in which data are generated and stored in a geo-distributed manner across vehicles and RSUs.

\subsection{M3T FedFMs: Blending FL and M3T FMs}
M3T FedFMs are gaining high traction in both ML and wireless/communication communities as one of the major next technologies~\cite{ren2025advances}. Nevertheless, despite the extensive literature
on conventional FL in vehicular networks and
the growing interest in federated foundation models,
the joint consideration of multi-modal sensing,
multiple downstream tasks, and modular adaptation
of pre-trained FMs across heterogeneous vehicular
edge nodes remains comparatively underexplored. The most related work is pFedLVM in~\cite{kou2024pfedlvm}, which addresses the challenges of deploying Large Vision Models (LVM) in federated settings by keeping the LVM backbone centralized and sharing latent features with vehicles, enabling personalized learning. 
 To our knowledge, pFedLVM is among the first to present a framework for  the implementation of M3T FedFMs over vehicular networks. 
\textit{Accordingly, our overarching goal is to elucidate
how the interplay between pre-trained FMs,
heterogeneous modality--task profiles of vehicles, modular
federated adaptation of FMs, and dynamic vehicular
environments introduces distinct architectural
and learning challenges. Building on this
perspective, we identify corresponding research
directions and provide a representative case
study on federated task onboarding to facilitate
further investigations of M3T FedFMs in
vehicular networks.}

\section{Learning Architecture of M3T FedFMs}
In a nutshell, M3T FedFMs enable the distributed training of M3T FMs across a set of decentralized devices (e.g., vehicles). Owing to the recency and ongoing evolution of this research direction, there is currently no unique learning architecture for M3T FMs, and, by extension, for M3T FedFMs. Subsequently, to ground our discussion, and informed by recent advances in this area~\cite{chen2024feddat,ren2025advances}, we delineate a modular architecture for M3T FMs by decomposing their design into three core components, and then discuss how this modularity extends to the M3T FedFM paradigm.

\begin{figure}
    \centering
    \includegraphics[width=0.99\linewidth]{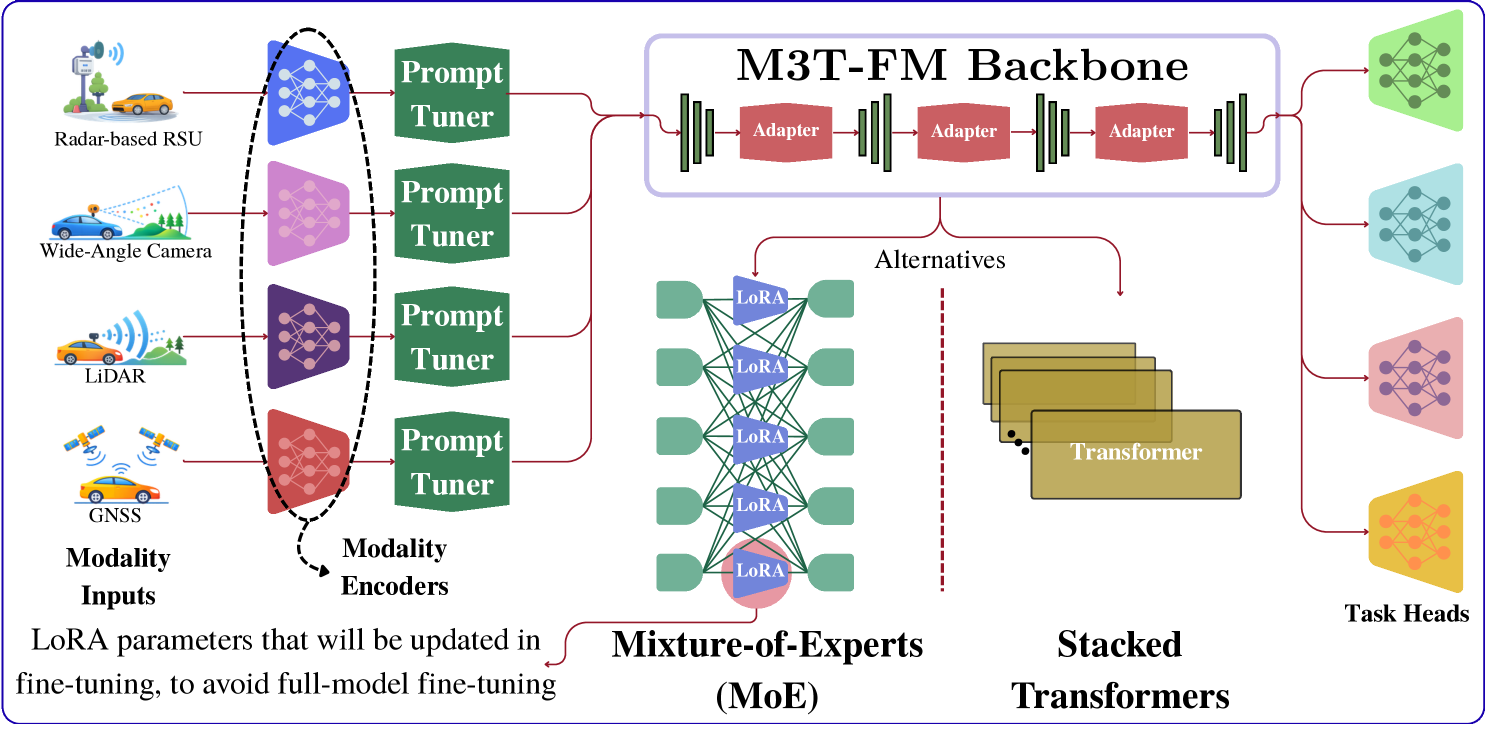}
    \vspace{-8mm}
    \caption{Schematic of the M3T FM architecture, consisting of modality encoders, a shared backbone (e.g., Mixture-of-Experts (MoE) or stacked transformers), and task heads. In addition, lightweight adaptation mechanisms, such as prompt tuning, low-rank adaptation (LoRA),  and adapter layers can be employed for parameter-efficient model fine-tuning.}
    \label{fig:system_figure}
\end{figure}

\subsection{Architecture of M3T FMs}

In this paper, we use the term M3T FM to refer
to an FM-based architecture that satisfies
three key characteristics: (i) it builds upon a
pre-trained FM with transferable knowledge;
(ii) it can process and integrate at least two
distinct input data modalities; and (iii) it
supports at least two distinct downstream learning
tasks through shared representations and
task-specific mechanisms. As illustrated in Fig.~\ref{fig:system_figure}, the architecture of an M3T FM can be decomposed into the following three components:

\textbf{1. Modality Encoders:} Modality encoders serve as the front-end of an M3T FM, with each encoder responsible for processing a specific input modality (e.g., LiDAR, wide-angle cameras, or GNSS signals). Specifically, each modality encoder transforms raw sensory inputs into a latent embedding representation, which captures the salient features of the corresponding modality and is subsequently passed to the model's \textit{backbone} for further processing.

\textbf{2. Backbone:} The model backbone is responsible for cross-modal alignment and fusion, producing unified latent representations that can be adapted to downstream tasks. This backbone can take a variety of architectural forms, two common choices of which are outlined below: 

\textbf{~~(i) Stacked Transformers (e.g., GPT, Gemini,~CLIP~\cite{rothman2024transformers}):} Stacked transformer backbone architectures employ multiple layers of self-attention and feed-forward networks to capture long-range dependencies and complex interactions across modalities.

\textbf{~~(ii) Mixture-of-Experts (e.g., DeepSeek~\cite{liu2024deepseek}):} Mixture-of-Experts (MoE) backbones decompose the model into a collection of specialized \textit{expert} subnetworks, with a gating mechanism dynamically routing inputs to a subset of experts. This design enables conditional computation, allowing different modalities, tasks, or input contexts to activate different experts.

\textbf{3. Task Heads:} Task heads are typically implemented as lightweight neural modules that map the embeddings produced by the model's backbone to concrete task-specific predictions, such as control commands or action/decision probabilities.

It is worth noting that the above components of an M3T FM
may not all originate from the same pre-training stage.
In particular, a pre-trained FM can serve
as the architectural and representational basis,
while additional modality-specific encoders, fusion
modules, and task-specific heads can be introduced
to extend its capabilities to new modalities and
downstream tasks.

\begin{figure*}
    \centering
    \includegraphics[width=0.99\linewidth]{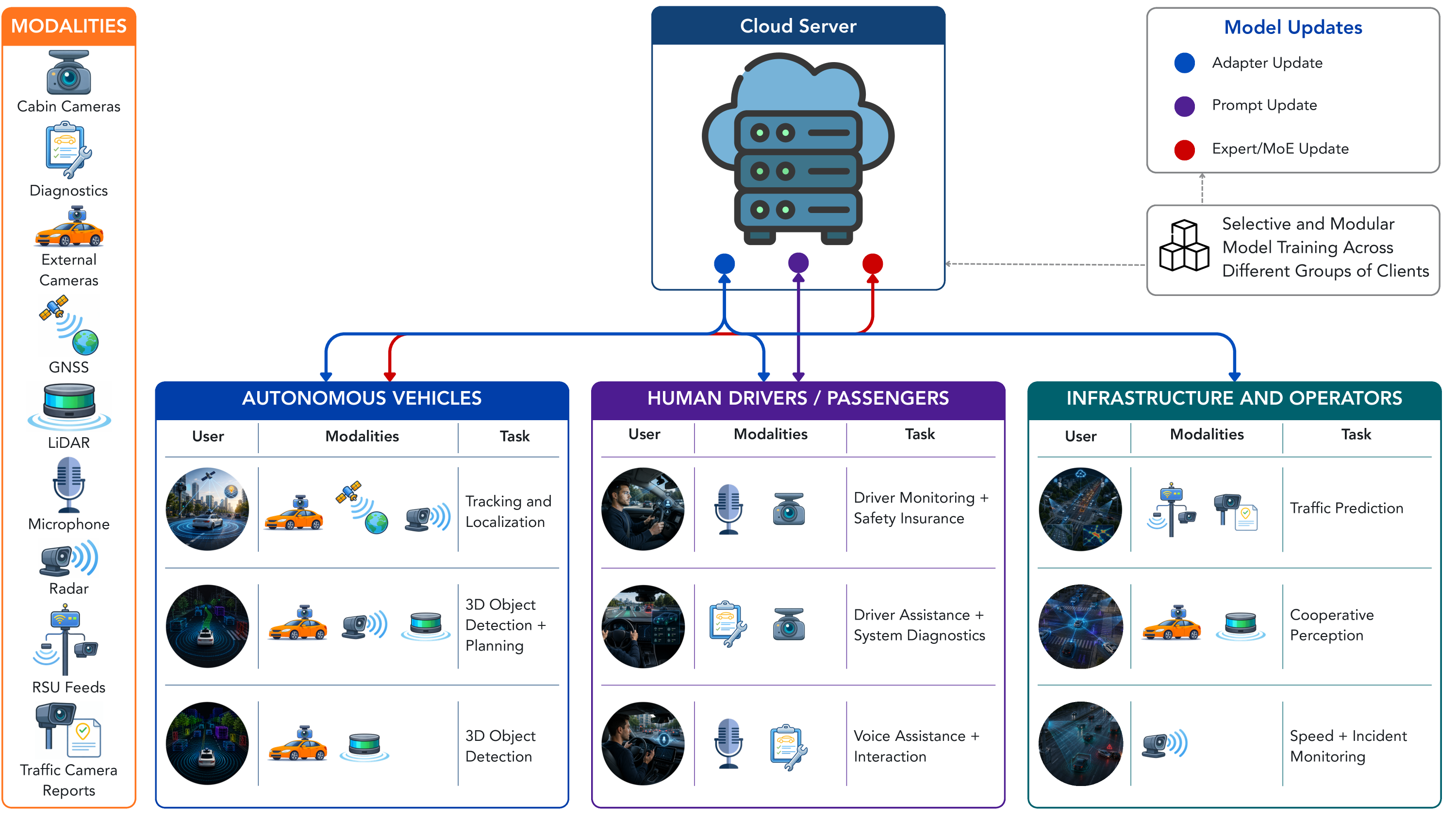}
    \vspace{-4mm}
    \caption{Schematic of the M3T FedFM architecture over a set of vehicular edge nodes (the orange box on the left collects the existing data modalities in the diagram). Each vehicular edge node performs local fine-tuning of its model using parameter-efficient adaptation techniques (e.g., prompt tuning, LoRA, or adapter layers) to train/fine-tune its local M3T FM and transmits the resulting model updates to a server. The server then aggregates the received model updates to refine the global model parameters and disseminates the updated global model back to the nodes for subsequent training rounds.}
    \label{fig:system_figure2}
\end{figure*}
\subsection{Architecture of M3T FedFMs}

Given that M3T FMs typically comprise billions/trillions of parameters, they commonly undergo an initial \textit{pre-training} phase, often conducted centrally on massive and diverse datasets, to acquire generalizable  knowledge. Unlike conventional FL, which commonly involves training task-specific models from scratch, M3T FedFMs leverage this pre-existing knowledge to enable distributed training/fine-tuning across geographically distributed devices using local datasets relevant to the downstream tasks of interest. This shifts the learning objective toward adapting pre-trained M3T FMs to heterogeneous modalities and downstream tasks while accounting for their substantial computational and memory requirements.


 
Specifically, in M3T FedFMs (see Fig.~\ref{fig:system_figure2}), full-model training is typically avoided, as it is often infeasible across resource-constrained clients. Instead, clients employ parameter-efficient fine-tuning  (PEFT) techniques to adapt selected model components/modules locally, and subsequently transmit their locally learned model updates (e.g., gradients or module parameters) to a coordinating server.
The server aggregates these updates and disseminates the refined parameters back to the clients for subsequent training rounds. 
While the aggregation process can still follow
conventional FL principles (e.g., FedAvg),
M3T FedFMs commonly aggregate selected
PEFT-enabled modules rather than the entire model.
Moreover, in heterogeneous multi-modal multi-task
settings, different clients may train distinct
subsets of these modules according to their
available modalities and downstream tasks,
necessitating selective aggregation of compatible
model updates.
In particular, representative PEFT techniques and their subsequent PEFT-enabled modules (see Fig.~\ref{fig:system_figure} for their placement within the local M3T FMs of nodes) are as follows:
\begin{itemize}[leftmargin=4mm]
    \item \textbf{Low-Rank Adaptation (LoRA):}  
   Low-Rank Adaptation (LoRA) injects trainable low-rank matrices into selected backbone layers; these matrices constitute the PEFT modules that are trained and aggregated across clients.
    
\item \textbf{Prompt Tuning:}  
Prompt tuning prepends a small set of trainable prompt embeddings to the backbone input; these embeddings constitute the PEFT modules that are trained and aggregated across clients.

\item \textbf{Adapter Tuning:}  
Adapter tuning inserts lightweight, trainable bottleneck layers between selected backbone layers; these layers constitute the PEFT modules that are trained and aggregated across clients.

\item \textbf{Mixture-of-Experts (MoE) Training:}  
MoE training updates a subset of expert subnetworks activated for specific inputs, modalities, or tasks; the parameters of these experts (and, when applicable, the gating network) constitute the PEFT modules that are trained and aggregated across clients.

\end{itemize}

\section{Use Cases of M3T FedFMs in Vehicular Networks}
In this section, we outline representative use cases of M3T FedFMs across key stakeholders in vehicular ecosystems, including autonomous vehicles, human drivers and passengers, and vehicular network operators. For each use case, we (i) identify the typical data modalities involved, (ii) describe the expected downstream tasks, and (iii) present an example that highlights the potential of M3T FedFMs in vehicular settings.

\subsection{Use Case 1: M3T FedFM-Powered Autonomous Vehicles}
M3T FedFMs across \textit{autonomous vehicles} (e.g., self-driving cars, buses, and delivery fleets) allow each vehicle to adapt an M3T FM to its local sensing conditions and operational objectives/tasks without sharing raw sensory data. In particular, autonomous vehicles are typically equipped with multi-modal sensing capabilities, including LiDAR, cameras, millimeter-wave radar, GNSS signals, inertial sensors, and vehicle-state telemetry. Leveraging these locally collected data modalities, vehicles pursue a diverse set of downstream objectives/tasks, such as localized perception (e.g., 3D object detection and tracking), prediction (e.g., trajectory and intent inference), planning and control, localization and mapping, and cooperative perception with nearby vehicles or roadside infrastructure. Within this context, M3T FedFMs facilitate collaborative, module-level fine-tuning of M3T FMs across vehicles. For example, in an M3T FedFM paradigm, vehicles observing different portions or viewpoints of the same environment can collaboratively fine-tune their M3T FMs using their own sensor data to improve their cooperative perception performance, such as detecting occluded objects or enhancing scene completeness.

\subsection{Use Case 2: M3T FedFMs for Human Drivers/Passengers}\label{sec:usecase-human}
M3T FedFMs enable the distributed fine-tuning of M3T FMs across vehicles to support intelligent, interactive services for human drivers and passengers, allowing each vehicle to adapt an M3T FM to user-specific behaviors, preferences, and contexts without sharing raw in-cabin or diagnostic data. In particular, modern vehicles collect a wide range of local data modalities, including cabin cameras, microphones, driver-monitoring sensors, dashboard signals, vehicle diagnostics, and contextual information from infotainment systems. Using these locally collected modalities, vehicles support downstream tasks such as driver monitoring, personalized assistance, voice- and vision-based interaction, comfort and safety optimization, and in-vehicle diagnostics. 
For example, when a driver encounters a dashboard warning or suspected mechanical issue, the driver can first capture an image of the warning indicator or the relevant vehicle component. The locally fine-tuned M3T FM then analyzes this visual input, along with available contextual signals (e.g., vehicle diagnostics or recent sensor readings), to provide an initial explanation and recommended actions. If needed, the driver can follow the model's guidance to capture additional images or provide further contextual input, allowing the model to iteratively refine the diagnosis and troubleshooting steps, thereby augmenting individual mechanical expertise and reducing reliance on rather expensive vehicle service centers.

\subsection{Use Case 3: M3T FedFMs for Vehicular Network Operators}
M3T FedFMs can unlock the distributed fine-tuning of M3T FMs across vehicular network operators, allowing transportation authorities, city planners, and mobility-as-a-service providers (e.g., ride-hailing services, on-demand shuttle operators, and shared-mobility fleet managers) to collaboratively train their M3T FMs using their region-specific traffic patterns and infrastructure conditions without centralizing raw data. In particular, vehicular network operators collect and access a diverse set of data modalities, including vehicle reports, RSUs' collected data, traffic cameras, environmental sensors, and infrastructure telemetry. Further, the expected downstream tasks include traffic flow prediction, road-condition monitoring, accident detection and reporting, fault and liability analysis (e.g., identifying contributing factors in collisions), intelligent traffic signal coordination (e.g., adaptive red-light scheduling), and proactive congestion mitigation. 
For example, operators in different regions can locally fine-tune their M3T FMs on their traffic and infrastructure data to improve incident detection and traffic control policies, where the model aggregation across these operators through the M3T FedFM paradigm enables the training of a global model that progressively improves its understanding of traffic dynamics and roadway conditions across the broader vehicular ecosystem.

Digital twin (DT) technology can further complement this
paradigm by maintaining virtual representations of
vehicles and road infrastructure, enabling simulation
and evaluation of traffic conditions and control
strategies. Inspired by recent advances in AI-enabled
DT systems~\cite{chen2025generative,chen2024revolution},
M3T FedFMs could leverage such virtual environments
to support collaborative model adaptation and evaluate
new traffic management policies under evolving
vehicular and network conditions.

\section{Challenges and Future Research Directions}
We next focus on the challenges associated with implementing M3T FedFMs in vehicular networks. To provide a more structured presentation, we organize our discussions around four key challenges, each paired with a set of research directions.

\subsection{
Challenge 1: M3T FedFMs under Unpredictable Vehicle Availability, Edge Handoffs, and Limited Computation/Communication Resources}

Vehicular networks exhibit highly heterogeneous and time-varying client/vehicle availability, which poses a fundamental challenge to the training and fine-tuning of M3T FedFMs. In practice, vehicles participate in model training under vastly different operational conditions: for example, an electric vehicle parked at a charging station or in a parking lot may have ample energy, stable connectivity, and idle compute resources, allowing it to engage in local model fine-tuning with minimal impact on its primary functions. In contrast, vehicles that are actively driving must balance model training with latency-critical perception, planning, and control tasks, often resulting in intermittent or constrained participation in learning.
Moreover, even when vehicles are capable of performing local training while in motion, they frequently traverse the coverage areas of multiple base stations, leading to frequent handoffs. As a result, the set of vehicles participating in each training round of M3T FedFM can vary over time and cannot be assumed to be stable or known a priori. These dynamics introduce training fragmentation, where model updates are generated by a continuously changing subset of vehicles with heterogeneous availability, connectivity, and resource profiles. Furthermore, although PEFT techniques reduce the
number of trainable parameters, local fine-tuning
of large-scale M3T FMs may still impose substantial
computational and memory demands. Moreover,
repeatedly exchanging adapter, LoRA, or expert
updates over multiple aggregation rounds can incur
considerable communication overhead. These
constraints become particularly pronounced under
limited wireless bandwidth, short connectivity
durations, and frequent vehicular handoffs.

Beyond their impact on empirical learning performance,
vehicle mobility and frequent handoffs can also
complicate the establishment of theoretical convergence
guarantees for M3T FedFMs. In particular, mobility-driven
client participation may introduce selection bias
(i.e., the systematic overrepresentation or underrepresentation
of vehicles with certain data characteristics in the model
training process), potentially causing the aggregated model
updates to deviate from the intended global learning objective.
Moreover, frequent handoffs may introduce delayed or
stale model updates in M3T FedFM settings,
while heterogeneous connectivity durations can lead to
varying numbers of local fine-tuning iterations
(e.g., stochastic gradient descent steps) across
vehicles. These factors can exacerbate client drift
(i.e., the divergence of locally trained model parameters
from the global model due to heterogeneous local updates)
and introduce additional errors in the global model updates,
potentially slowing convergence or weakening convergence
guarantees, particularly under non-independent and identically distributed (non-IID) vehicular data
distributions.

Consequently, M3T FedFMs in vehicular networks operate in an inherently on-the-fly learning regime, in which participant selection, training duration, and communication opportunities must adapt dynamically to evolving mobility and network conditions.
Addressing this challenge calls for the development of availability- and mobility-aware M3T FedFM frameworks that explicitly account for unpredictable vehicle participation and edge handoffs. In this context, promising research directions include (\textit{i) joint communication-computation-learning
optimization for M3T FedFMs,} focusing on dynamically
selecting/scheduling participating vehicles, trainable PEFT
modules, local fine-tuning durations, and aggregation
frequencies according to available wireless bandwidth,
computational and memory resources, energy constraints,
and predicted connectivity durations, while balancing
resource consumption against model convergence and
downstream task performance; \textit{(ii) mobility-aware convergence analysis for M3T FedFMs},
focusing on characterizing the impact of handoff-induced
variations in client participation, model update staleness,
and heterogeneous local fine-tuning durations on convergence
rates and theoretical convergence guarantees (e.g.,
convergence to stationary points for non-convex objectives),
while identifying sufficient conditions (e.g., bounded
update delays, adequate long-term client participation,
and controlled local training durations) under which
such convergence guarantees can be established.

\subsection{Challenge 2: M3T FedFMs under Vehicular Task and Hardware Diversity}

Vehicular networks are inherently characterized by task and hardware diversity across participating edge nodes, which poses a major challenge to the training and fine-tuning of M3T FedFMs. Specifically, different edge nodes within the vehicular ecosystem, including autonomous vehicles, human-driven vehicles, RSUs, and other sensing or data-collection infrastructure, are tasked with distinct learning objectives. For instance, vehicles may prioritize perception, prediction, and planning tasks, while RSUs may focus on traffic monitoring, incident detection, or infrastructure-aware sensing. As a result, participating nodes often contribute data and model updates corresponding to heterogeneous and partially overlapping task sets.
This challenge is further exacerbated by hardware and sensing heterogeneity across these edge nodes: vehicles and infrastructure units are equipped with different sensing suites, ranging from cameras, LiDAR, and radar to microphones, environmental sensors, and vehicle diagnostics, leading to variation in the available data modalities across clients. 

Consequently, M3T FedFMs must be trained under unbalanced modality-task availability, where certain modalities or tasks are presented at some nodes and absent at others. 
Addressing this challenge calls for the development of task- and modality-aware M3T FedFM frameworks. Promising research directions include 
\textit{(i) selective and modular model training
and aggregation strategies for M3T FedFMs,}
in which only task-relevant or modality-specific
submodules of the local M3T FedFMs are updated,
and aggregation is performed selectively over
compatible modules across clients based on
their local modality--task profiles; \textit{(ii) adaptive client scheduling mechanisms for M3T FedFMs under unbalanced modality–task availability}, which dynamically select participants in model training and their training configurations according to their available modalities and tasks; such scheduling mechanisms must be revisited and tailored to the underlying PEFT approach used for model fine-tuning.
\begin{figure*}[!h]
    \centering
    \includegraphics[width=\linewidth]{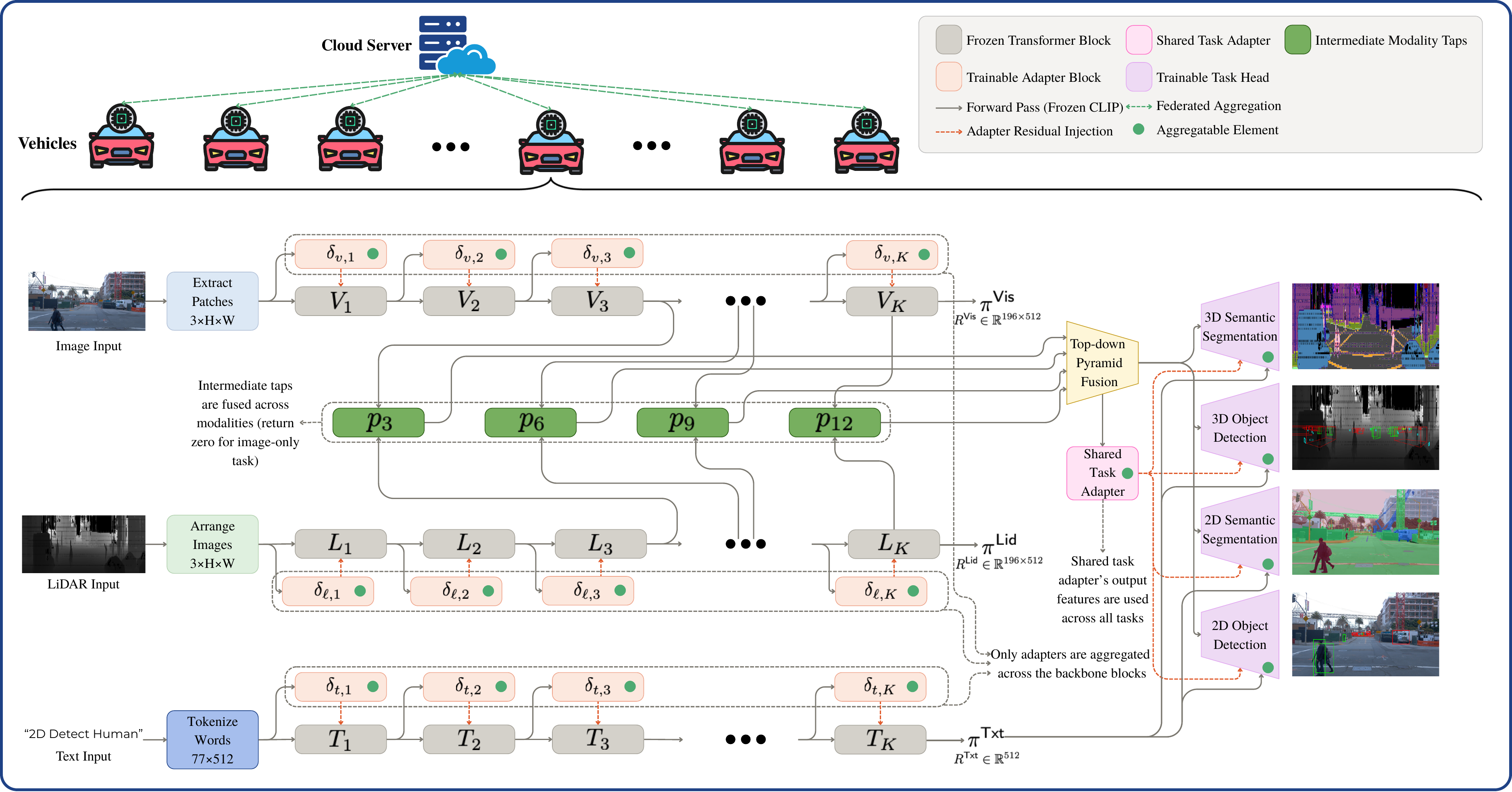}
    \vspace{-7mm}
    \caption{Architecture of the proposed M3T FedFM framework and the corresponding federated adaptation process.
    The pre-trained CLIP/SigLIP2 vision and text encoders provide
    the basis of the local M3T FM at each client/vehicle, while the LiDAR
    encoder, multi-modal fusion module, shared task adapter,
    and task-specific prediction heads extend the architecture
    to support multi-modal multi-task learning. The LiDAR encoder is initialized using the pre-trained
    CLIP/SigLIP2 vision encoder parameters, whereas the additional
    fusion and task-adaptation modules are introduced
    to accommodate the target vehicular perception tasks. More specifically,
    the model consists of three modality-specific transformer encoder branches devoted to processing vision/image, LiDAR, and text modalities, denoted by $\{V_k\}_{k=1}^{K}$, $\{L_k\}_{k=1}^{K}$, and $\{T_k\}_{k=1}^{K}$, respectively (similar to CLIP/SigLIP2, the total number of transformer blocks is $K=12$). As in SigLIP2, intermediate modality taps (dark green blocks) are taken from the $3$rd, $6$th, $9$th, and $12$th blocks of the transformer backbone for image and LiDAR modalities to account for different levels of dense information stored in each transformer's output; i.e., earlier transformer layer outputs contain more dense (pixel-level) information while later transformer layer outputs contain more abstract (patch-level) information. Each modality's intermediate taps are then reassembled by turning the $14\times14$ sized outputs of the transformer blocks $3$, $6$, $9$, and $12$ into the dimensions $56^2$, $28^2$, $14^2$ and $7^2$, creating an image pyramid. Each corresponding intermediate modality tap is then fused across modalities and the four taps are then fused in a top-down manner to form a multi-modal image-LiDAR representation.
    The image, LiDAR, and text inputs are first transformed into patch/token embeddings of size $3 \times H \times W$, $3 \times H \times W$, and $77 \times 512$ for the image, LiDAR, and text modalities, respectively, and subsequently propagated through their corresponding transformer backbones. Here, 
    $H$ and $W$ correspond to the height and width of each image/LiDAR patch. To enable parameter-efficient fine-tuning, lightweight trainable adapters, denoted by $\{\delta_{v,k}\}_{k=1}^{K}$, $\{\delta_{\ell,k}\}_{k=1}^{K}$, and $\{\delta_{t,k}\}_{k=1}^{K}$, are inserted into the frozen backbone blocks via residual adapter injections. The resulting modality-specific semantic embeddings are denoted by $\pi^{\mathrm{Vis}}\in\mathbb{R}^{196\times 512}$, $\pi^{\mathrm{Lid}}\in\mathbb{R}^{196\times 512}$, and $\pi^{\mathrm{Txt}}\in\mathbb{R}^{512}$, corresponding to the latent feature representations extracted from the vision, LiDAR, and text branches, respectively, and are used for modality alignment analysis. The fused representation of image and LiDAR modalities in the top-down pyramid fusion is processed by a shared task adapter that captures transferable cross-task knowledge across tasks. The resulting shared representation is subsequently delivered to multiple task-specific heads supporting the downstream tasks. Ultimately, at the end of each task head, a cosine similarity operation is performed on the token embeddings from the text modality via dot product. Aggregatable parameters including the modality adapters, the shared task adapter, and the task heads are shown with green circles.}
    \label{fig:local-ml-model}
\end{figure*}

\subsection{Challenge 3: M3T FedFMs under Time-Varying Vehicular Data}
Vehicular networks operate in environments where both data modalities and learning tasks evolve/drift over time, posing a challenge to the training and deployment of M3T FedFMs. In practice, the sensing conditions experienced by vehicles and roadside infrastructure can change due to factors such as traffic density, weather, time of day, road topology, and hardware usage patterns. As a result, the availability, quality, and relevance of different data modalities (e.g., camera, LiDAR, radar, or contextual signals) may drift over time. Similarly, the importance and definition of downstream tasks can shift, as vehicles transition between operational contexts, such as urban driving, highway cruising, construction zones, or adverse weather conditions, each emphasizing different perception, prediction, or control objectives.
These temporal variations induce modality and task drift across edge nodes in vehicular ecosystems, leading to non-stationary learning environments in which data distributions and task priorities change continuously. 

Consequently, M3T FedFMs must operate under evolving modality–task profiles in vehicular networks. In this context, promising research directions include
\textit{(i) continual learning for M3T FedFMs under modality and task drift,} focusing on developing PEFT-driven continual learning mechanisms that allow M3T FedFMs to incorporate new information as modalities and tasks evolve, while mitigating catastrophic forgetting across previously learned modalities and tasks across the vehicular edge nodes; 
\textit{(ii) instantaneous model tracking for M3T FedFMs under time-varying modality–task profiles,} focusing on maintaining the best-performing PEFT modules for the \emph{current} modality-task profile across vehicular edge nodes, prioritizing rapid PEFT module adaptation to recent data over long-term knowledge retention.

\subsection{Challenge 4: Privacy Leakage through Model Updates}

Although M3T FedFMs eliminate the need to centralize raw vehicular data, this property alone does not guarantee privacy, as exchanged model parameters/gradients may still leak information about local training datasets through gradient inversion, membership inference, or other reconstruction attacks. Such risks are particularly concerning in vehicular environments, where locally collected data modalities may contain sensitive information about vehicle locations, driving behaviors, and passengers. Moreover, PEFT methods do not inherently eliminate such vulnerabilities, since shared LoRA parameters, adapters, or other trainable modules in M3T FedFMs may still encode information about private data.

Addressing this challenge calls for the development of privacy-aware M3T FedFM frameworks that integrate mechanisms such as differential privacy, secure aggregation, and cryptographic protection of model parameters/gradients. In this context, promising research directions include
\textit{(i) designing privacy-preserving aggregation mechanisms tailored to heterogeneous PEFT modules}, where privacy protection mechanisms are adapted to the distinct structures and sensitivities of the trainable modules  (e.g., LoRA matrices, prompt embeddings, and adapters) exchanged across vehicular edge nodes in M3T FedFMs, while accounting for the varying privacy requirements of different data modalities and tasks;
\textit{(ii) jointly optimizing privacy protection, model utility, and communication overhead}, focusing on dynamically adapting privacy protection levels and aggregation strategies (e.g., adjusting differential privacy noise levels, selectively applying secure aggregation to sensitive PEFT modules, and adapting client participation based on privacy and resource constraints) according to the available communication and computational resources of vehicular edge nodes, while limiting the degradation in downstream task performance.

\section{Case Study}
We next perform a case study on a representative challenge in M3T FedFMs: the ability to incorporate new tasks into an already-trained model without retraining from scratch or disrupting existing capabilities. This challenge, known as \textit{task onboarding} or \textit{continual task arrival}, is acute in vehicular networks, where a vehicle fleet may have been trained on a set of perception capabilities and must integrate new tasks over time as operational requirements evolve. 

The details of our implementations, including all source codes, are provided in our GitHub repository:
\href{https://github.com/KasraBorazjani/vehicular-fedfm}{https://github.com/KasraBorazjani/vehicular-fedfm}

\begin{figure*}[t]
    \centering
    \includegraphics[trim={0 0 0 1cm}, clip, width=\linewidth]{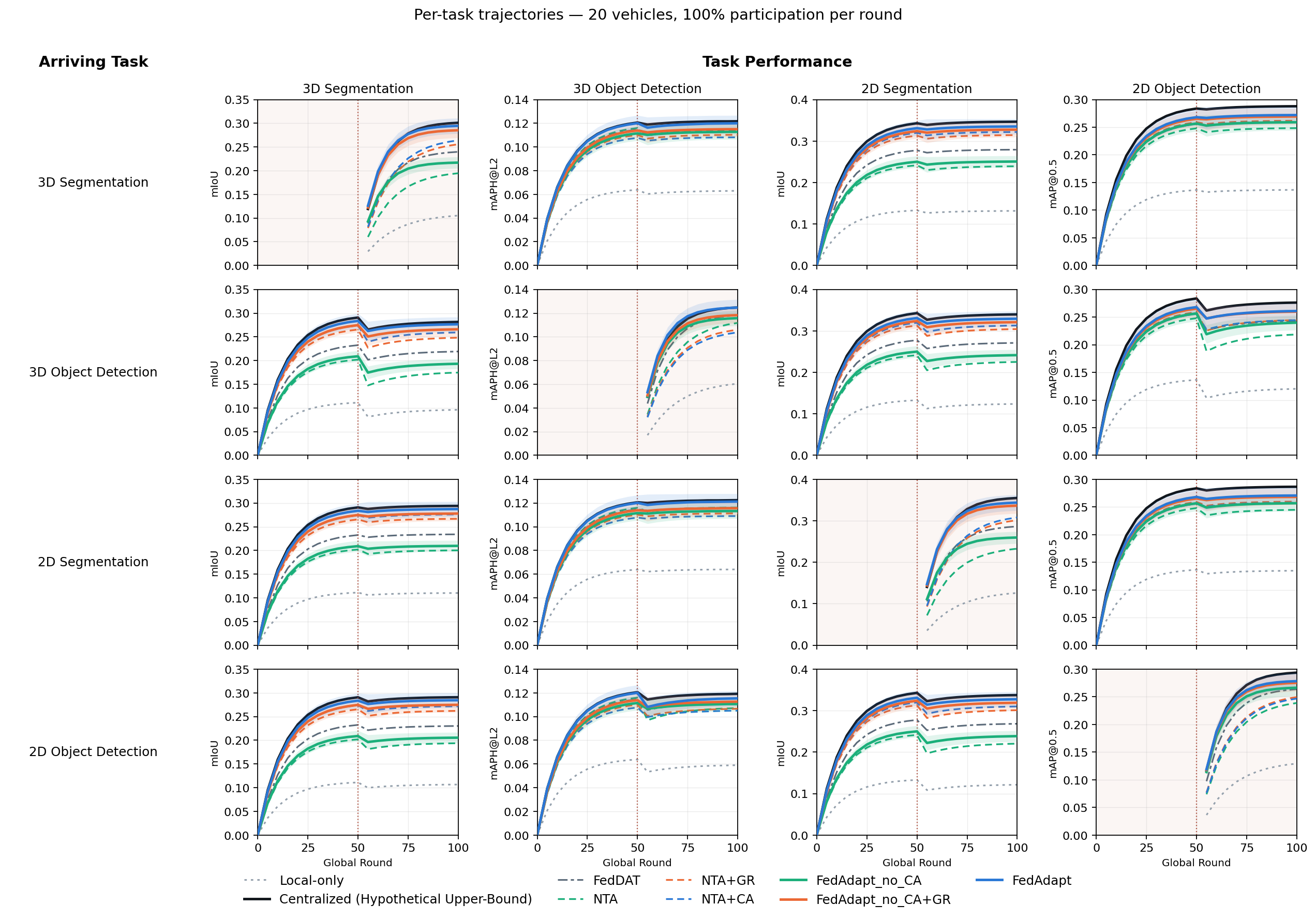}
    \vspace{-8mm}
    \caption{Task onboarding performance over the M3T FedFM global rounds for the arriving task (diagonal subfigures) and pre-trained tasks (off-diagonal subfigures) under each task-arrival scenario, as indicated on the left side of each row. The vertical dotted line at global round 50 indicates the arrival of the new task being onboarded in each scenario. FedAdapt is the best-performing method for all arriving tasks, followed by FedAdapt$\_$no$\_$CA+GR, demonstrating the benefit of our proposed shared task adapter when combined with gradient conflict resolution.}
    \label{fig:performance-rounds}
    \vspace{-1mm}
\end{figure*}

\subsection{Dataset and Tasks}

We implement an M3T FedFM over an urban vehicular scenario comprising 20 vehicles/clients based on the real-world Waymo Open Dataset's perception segment (available online at \href{https://waymo.com/open/data/perception/}{https://waymo.com/open/data/perception/}).
The dataset contains data gathered from cameras deployed on cars across San Francisco, Mountain View, Los Angeles, Detroit, Seattle, and Phoenix. 
Each segment comprises a 20-second driving clip sampled at approximately 10\,Hz, captured by five surround-view cameras providing full 360° coverage and one mid-range spinning LiDAR sensor. We use the calibration provided in the dataset for each segment to align each range image with the camera POVs to create multi-modal image-LiDAR training data pairs. We dedicate four dataset segments (three for training and one for validation) per-client. Since the segments are collected in different cities, weather/lighting conditions, and object class frequencies, the data across clients becomes inherently non-IID.

We train our M3T FedFM on four perception tasks spanning both camera and LiDAR  modalities alongside an augmented text modality:
(i) \textit{3D Semantic Segmentation,} where the input modalities are image, LiDAR, and text, while the output is the per-point semantic class label in the LiDAR point cloud across 22 object categories. The performance metric used for this task is mean Intersection over Union (mIoU).
(ii) \textit{3D Object Detection,} where the input modalities are image, LiDAR, and text, while the output is a predicted set of box center coordinates ($x$, $y$, $z$), box dimensions in each direction ($dx$, $dy$, $dz$) and a heading angle for each object in the scene. The performance metric used for this task is mean Average Precision weighted by Heading accuracy at difficulty level L2 (mAPH/L2).
(iii) \textit{2D Semantic Segmentation,} where the input modalities are image and text, while the output is the per-pixel semantic class for 28 semantic classes. The performance metric used for this task is mean Intersection over Union (mIoU).
(iv) \textit{2D Object Detection,} where the input modalities are image and text, while the output is 2D axis-aligned bounding boxes for vehicles, pedestrians, cyclists, and signs in each of the five camera images. The performance metric used for this task is mean Average Precision at IoU threshold 0.5 (mAP@0.5). Further, the learning rate for task heads and adapter group was selected as $1e-3$, mini-batch size is 16, number of local gradient optimization steps before each aggregation is 25, local model optimizer was chosen as ``adamw" with weight decay set to $1e-4$.

\subsection{Local M3T FM Architecture and Task Onboarding Setup}
We consider SigLIP2~\cite{tschannen2025siglip}, a CLIP~\cite{rothman2024transformers} variation with its architecture adapted for dense visual tasks such as segmentation and detection, as the underlying M3T FM model deployed across the clients. Since this model is originally designed for handling text-image modalities and performing a single task, we make the following advancements to the model: (i) we add a new backbone for LiDAR, where the transformers are initialized with those of the image branch (the encoder branches for vision/image, LiDAR, and text are denoted by $V_{\cdot}$, $L_{\cdot}$, and $T_{\cdot}$ in Fig.~\ref{fig:local-ml-model}), and (ii) we supplement the model with a series of task heads, each consisting of a task-specific architecture with LSeg-style decoders for segmentation and CenterHead-style decoders for detection (see our GitHub for the exact implementation).

To handle the task onboarding in the M3T FedFM setup, we design and add a \textit{shared task adapter} to the model, which captures cross-task representations (see Fig.~\ref{fig:local-ml-model}). The rationale behind this design relies on the capability of the shared task adapter to contain information from the pre-trained tasks and act as an anchor/teacher for the arriving task. Thus, we will have task-related information that is re-adapted with each arriving task and usable for all arriving tasks. This also lightens the burden on the model backbone's parameters that are intended to extract multi-modal features from the input rather than adapting to the tasks present in the system. 
It is worth emphasizing that our contribution lies not in
introducing new PEFT or gradient optimization techniques,
but rather in developing a task-onboarding-oriented M3T FedFM
architecture for vehicular networks, a setting that remains
relatively underexplored. In particular, we extend the
pre-trained SigLIP2 architecture with a LiDAR encoder branch
and introduce a shared task adapter to facilitate knowledge
transfer from previously learned tasks to newly arriving ones
in an M3T FedFM setting. Our ultimate method, called \textit{FedAdapt}, adopts an adapter-based PEFT technique (adapters across the image/vision, LiDAR, and text encoder branches are denoted by $\delta_{v,\cdot}$, $\delta_{\ell,\cdot}$, and $\delta_{t,\cdot}$ in Fig.~\ref{fig:local-ml-model}) and integrates this shared task adapter along with local application of Conflict-Averse Gradient Descent (CAGrad~\cite{liu2021conflict}) across clients to account for disproportionate task gradient conflicts during the backpropagation.

For each reported result, one task is designated as the \textit{arriving task} while the remaining three tasks are treated as \textit{pre-trained tasks} whose parameters are loaded directly from M3T FedFM prior to the new task arrival and only their task heads are fine-tuned after the new task arrival. When the arriving task is onboarded, the following training protocol is applied: the task head corresponding to the arriving task is initialized from scratch and trained; the shared task adapter (where applicable) and all modality-specific adapters (visual, LiDAR, and text) are fine-tuned to accommodate the new task; and other parameters, including the frozen SigLIP2 backbone layers, remain unchanged throughout. 





\textbf{Baselines:}
We compare the performances of our method, \textit{FedAdapt}, to the following baselines (across all methods, FedAvg~\cite{mcmahan2017communication} is used for model aggregation):

\begin{itemize}[leftmargin=3mm]
    \item \textbf{No\_Task\_Adapter (NTA):} Each client trains its local model with modality-specific adapters (visual, LiDAR, and text) but \textit{without} a shared task adapter. 

    \item \textbf{NTA+GR:} This baseline combines NTA with a local-level implementation of GradNorm gradient balancing~\cite{chen2018gradnorm}, but \textit{without} using the shared task adapter.  Our local implementation of GradNorm on each client dynamically reweights per-task loss gradients to enforce balanced task learning rates, using the ratio of each task's current loss to its initial loss as a proxy for relative training speed. 

    \item \textbf{NTA+CA:} This baseline combines NTA  with CAGrad~\cite{liu2021conflict}, but \textit{without} using the shared task adapter. 

    \item \textbf{FedAdapt\_no\_CA:} The baseline takes the CAGrad component out of our proposed model training pipeline.


    \item \textbf{FedAdapt\_no\_CA+GR:} This baseline replaces CAGrad with GradNorm~\cite{chen2018gradnorm} in our original model training pipeline.

    \item \textbf{Centralized:} This baseline provides a hypothetical upper-bound for the performance of M3T FedFM where limitations such as data heterogeneity and model bias would not exist.

    \item \textbf{Local-only:} This baseline eliminates federated aggregation by engaging the clients solely on local training on their local datasets.

    \item \textbf{FedDAT~\cite{chen2024feddat}:} This baseline extends an existing M3T FedFM training method that does not propose shared task adapters or gradient balancing to an onboarding scenario for tasks that are different in their output domain.


\end{itemize}

\subsection{Discussion of the Results}
\label{sec:results}

\begin{figure*}[htbp]
    \centering
    \includegraphics[trim={0 0 0 1cm}, clip, width=\linewidth]{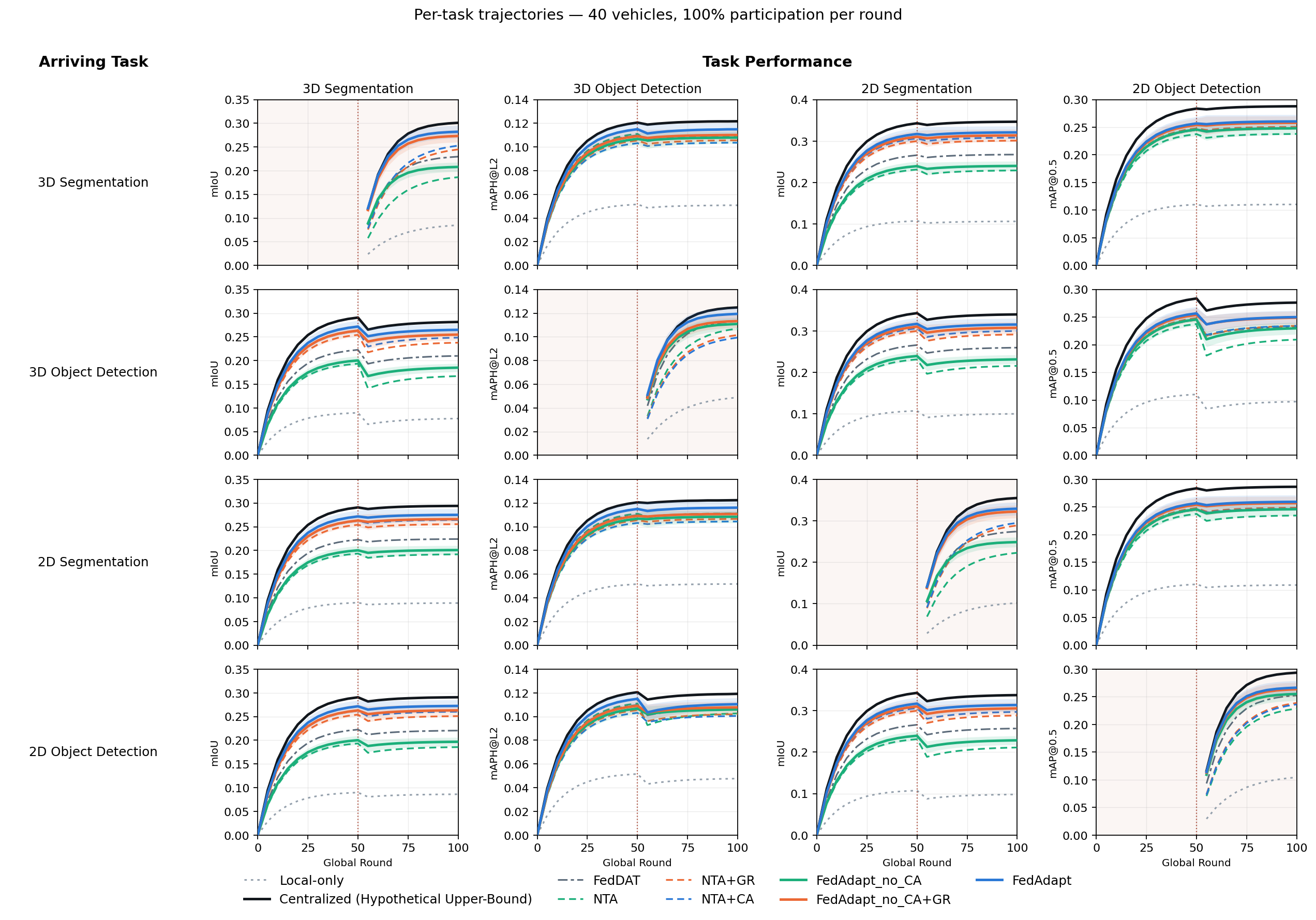}
    \vspace{-9mm}
    \caption{$40$-client task onboarding performance over the M3T FedFM global rounds for the arriving task (diagonal subfigures) and pre-trained tasks (off-diagonal subfigures) under each task-arrival scenario, as indicated on the left side of each row. The vertical dotted line at global round 50 indicates the arrival of the new task being onboarded in each scenario. FedAdapt is the best-performing method for all arriving tasks, followed by FedAdapt$\_$no$\_$CA+GR, demonstrating the benefit of our proposed shared task adapter when paired with a gradient conflict resolution method upon having a larger number of clients.}
    \label{fig:40-clients}
\end{figure*}

We present the results in Fig.~\ref{fig:performance-rounds}, which depicts the task onboarding performance over the model aggregation rounds for each arriving (diagonal sub-figures) and pre-trained (non-diagonal sub-figures) task under each arriving task, shown by the titles on the left of each row.
The most consistent finding is that the presence of our proposed shared task adapter provides a clear and consistent performance benefit compared to baselines that did not use a shared task adapter and gradient balancing combination (i.e., FedDAT and NTA baselines). FedAdapt exhibits the best performance across all federated baselines (excluding the hypothetical centralized baseline) for both pre-trained and arriving tasks, confirming that our proposed shared task adapter serves as an effective knowledge transfer mechanism in the task onboarding setting in M3T FedFMs when equipped with CAGrad. The presence of FedAdapt\_no\_CA+GR as the second-best-performing method (excluding the hypothetical centralized baseline) for both the arriving and the pre-trained tasks under all settings shows how the choice of gradient balancing method can affect the converged performance.
The local-only baseline performing the poorest in all scenarios indicates the importance of performing a federated global aggregation as the models will be overfit/biased on local data and lose performance.
It can also be observed that pre-trained tasks in Fig.~\ref{fig:performance-rounds} face a decline in performance in the first global round after the onboarding of the arriving task. This performance is recovered as the global rounds increase and FedAdapt is shown to be able to result in the best recovered performance after onboarding.
Additionally, we increase the number of clients to $40$ to evaluate FedAdapt in a more densely populated vehicular network. The results for this setting are presented in Fig.~\ref{fig:40-clients} and are consistent with those in Fig.~\ref{fig:performance-rounds}, further demonstrating that FedAdapt consistently outperforms the considered baselines as the number of clients increases.

\subsection{Further Results}
We also perform two additional studies which include the following:

\begin{enumerate}
    \item We change the client participation ratio in federated aggregation to consider the effects of a non-ideal communication scenario with rapid handoffs and high mobility. Experimenting with participation ratios of $\{25\%, 50\%, 75\%, 100\%\}$, we reached the results presented in Table~\ref{tab:participation}-(a) where it is shown that FedAdapt achieves the best performance under all values of participation ratio.
    \item We analyze the effect of adding a shared task adapter on preserving modality feature alignment through calculating a centered kernel alignment (CKA)~\cite{kornblith2019similarity} between the latent feature representations of the image and LiDAR modalities ($\pi^{\mathrm{Vis}}$ and $\pi^{\mathrm{Lid}}$). To isolate the effect of the shared task adapter, we compare the ratio of CKA after the task onboarding has taken place and its calculated value before the arriving task had arrived under the NTA and FedAdapt\_no\_CA settings where the only difference is the presence of a shared adapter (more details are given in the caption of Table~\ref{tab:alignment_ratio}). Our findings in Table~\ref{tab:alignment_ratio}-(b) show that FedAdapt\_no\_CA (i.e., our proposed shared task adapter) has been able to better preserve the modality feature alignment after the task onboarding.
\end{enumerate}

\begin{table*}[!h]
\centering
\caption{Robustness and alignment analysis of task onboarding (measured at global round $100$). (a)~Effect of client participation: per-task performance after onboarding each arriving task with $20$ vehicles under participation rates of $100\%$, $75\%$, $50\%$, and $25\%$.  Results are presented as mean $\pm$ standard deviation across three seeds. Centralized training does not depend on the participation rate or the number of clients, and local-only training does not depend on the participation rate, so they are reported once where applicable. In each row group, the best federated result is shown in bold. (b)~Modality feature alignment after onboarding each arriving task, given as $\mathrm{CKA}(\pi^{\mathrm{Vis}}, \pi^{\mathrm{Lid}})$ of the onboarded model divided by that of the pre-trained model; values closer to $1$ indicate better-preserved alignment.}
\label{tab:participation}
\label{tab:seeds}
\label{tab:alignment_ratio}
{\fontsize{4pt}{4.8pt}\selectfont
\renewcommand{\arraystretch}{1.5}
\setlength{\tabcolsep}{0.8pt}
\setlength{\aboverulesep}{0pt}
\setlength{\belowrulesep}{0pt}
\newcommand{\unc}[1]{\mbox{\fontsize{2.5pt}{3pt}\selectfont$\pm\,#1$}}
\newcommand{\uncbold}[1]{\mbox{\fontsize{2.5pt}{3pt}\selectfont$\boldsymbol{\pm}\,#1$}}
\newcommand{\stack}[2]{{\renewcommand{\arraystretch}{1}\begin{tabular}{@{}c@{}}#1\\#2\end{tabular}}}
\resizebox{\textwidth}{!}{%
\begin{NiceTabular}{|c||*{17}{c|}}
\CodeBefore
  \rectanglecolor{darkyellow}{1-1}{3-18}    
  \rectanglecolor{salmon}{4-1}{10-18}      
  \rectanglecolor{darksalmon}{11-1}{17-18} 
  \rectanglecolor{salmon}{18-1}{24-18}     
  \rectanglecolor{darksalmon}{25-1}{31-18} 
  \rectanglecolor{salmon}{32-1}{33-18}     
  \rectanglecolor{white}{34-1}{34-18}       
  \rectanglecolor{darkyellow}{35-1}{36-18}  
  \rectanglecolor{salmon}{37-1}{38-18}     
\Body
\toprule
\Block{1-18}{\textbf{(a) Effect of Client Participation Rate (20 Vehicles)}} & & & & & & & & & & & & & & & & & \\
\midrule
\Block{2-1}{\textbf{Part.}\\\textbf{Rate}} & \Block{2-1}{\textbf{Method}} & \Block{1-4}{\textbf{3D Seg. Arrives}} & & & & \Block{1-4}{\textbf{3D Det. Arrives}} & & & & \Block{1-4}{\textbf{2D Seg. Arrives}} & & & & \Block{1-4}{\textbf{2D Det. Arrives}} & & & \\
\cline{3-18}
 & & \textbf{3D Seg.} & \textbf{3D Det.} & \textbf{2D Seg.} & \textbf{2D Det.} & \textbf{3D Seg.} & \textbf{3D Det.} & \textbf{2D Seg.} & \textbf{2D Det.} & \textbf{3D Seg.} & \textbf{3D Det.} & \textbf{2D Seg.} & \textbf{2D Det.} & \textbf{3D Seg.} & \textbf{3D Det.} & \textbf{2D Seg.} & \textbf{2D Det.} \\
\midrule
\Block{7-1}{\textbf{100\%}} & NTA & $0.195$\unc{0.005} & $0.114$\unc{0.001} & $0.239$\unc{0.010} & $0.248$\unc{0.005} & $0.175$\unc{0.004} & $0.112$\unc{0.002} & $0.225$\unc{0.008} & $0.219$\unc{0.001} & $0.200$\unc{0.007} & $0.116$\unc{0.002} & $0.233$\unc{0.009} & $0.245$\unc{0.004} & $0.194$\unc{0.005} & $0.107$\unc{0.003} & $0.221$\unc{0.006} & $0.239$\unc{0.004} \\
 & NTA+GR & $0.256$\unc{0.003} & $0.110$\unc{0.002} & $0.314$\unc{0.008} & $0.262$\unc{0.002} & $0.248$\unc{0.006} & $0.106$\unc{0.003} & $0.304$\unc{0.007} & $0.245$\unc{0.003} & $0.267$\unc{0.009} & $0.111$\unc{0.002} & $0.302$\unc{0.008} & $0.260$\unc{0.002} & $0.262$\unc{0.008} & $0.107$\unc{0.003} & $0.301$\unc{0.006} & $0.250$\unc{0.007} \\
 & NTA+CA & $0.264$\unc{0.002} & $0.108$\unc{0.002} & $0.322$\unc{0.002} & $0.259$\unc{0.002} & $0.259$\unc{0.001} & $0.104$\unc{0.002} & $0.313$\unc{0.009} & $0.245$\unc{0.002} & $0.276$\unc{0.002} & $0.109$\unc{0.002} & $0.308$\unc{0.001} & $0.258$\unc{0.001} & $0.272$\unc{0.005} & $0.105$\unc{0.003} & $0.310$\unc{0.006} & $0.247$\unc{0.002} \\
 & FedDAT & $0.240$\unc{0.003} & $0.115$\unc{0.002} & $0.279$\unc{0.006} & $0.259$\unc{0.003} & $0.219$\unc{0.004} & $0.118$\unc{0.002} & $0.271$\unc{0.003} & $0.243$\unc{0.004} & $0.234$\unc{0.001} & $0.116$\unc{0.001} & $0.286$\unc{0.008} & $0.257$\unc{0.002} & $0.230$\unc{0.002} & $0.112$\unc{0.001} & $0.268$\unc{0.006} & $0.264$\unc{0.003} \\
 & FedAdapt\_no\_CA & $0.217$\unc{0.003} & $0.113$\unc{0.002} & $0.251$\unc{0.005} & $0.259$\unc{0.001} & $0.193$\unc{0.003} & $0.116$\unc{0.002} & $0.242$\unc{0.007} & $0.240$\unc{0.002} & $0.210$\unc{0.006} & $0.113$\unc{0.002} & $0.260$\unc{0.005} & $0.257$\unc{0.008} & $0.206$\unc{0.008} & $0.111$\unc{0.001} & $0.239$\unc{0.010} & $0.267$\unc{0.005} \\
 & FedAdapt\_no\_CA+GR & $0.285$\unc{0.004} & $0.115$\unc{0.002} & $0.328$\unc{0.007} & $0.269$\unc{0.002} & $0.266$\unc{0.009} & $0.118$\unc{0.001} & $0.321$\unc{0.005} & $0.260$\unc{0.001} & $0.278$\unc{0.002} & $0.116$\unc{0.002} & $0.337$\unc{0.001} & $0.268$\unc{0.001} & $0.275$\unc{0.006} & $0.113$\unc{0.002} & $0.319$\unc{0.003} & $0.275$\unc{0.003} \\
 & FedAdapt & {\boldmath$0.295$}\uncbold{0.008} & {\boldmath$0.120$}\uncbold{0.001} & {\boldmath$0.335$}\uncbold{0.009} & {\boldmath$0.272$}\uncbold{0.003} & {\boldmath$0.277$}\uncbold{0.005} & {\boldmath$0.125$}\uncbold{0.002} & {\boldmath$0.329$}\uncbold{0.006} & {\boldmath$0.261$}\uncbold{0.002} & {\boldmath$0.287$}\uncbold{0.002} & {\boldmath$0.121$}\uncbold{0.002} & {\boldmath$0.344$}\uncbold{0.005} & {\boldmath$0.271$}\uncbold{0.001} & {\boldmath$0.285$}\uncbold{0.001} & {\boldmath$0.115$}\uncbold{0.002} & {\boldmath$0.328$}\uncbold{0.010} & {\boldmath$0.278$}\uncbold{0.005} \\
\hline
\Block{7-1}{\textbf{75\%}} & NTA & $0.191$\unc{0.003} & $0.112$\unc{0.002} & $0.234$\unc{0.003} & $0.244$\unc{0.004} & $0.170$\unc{0.005} & $0.110$\unc{0.002} & $0.220$\unc{0.005} & $0.213$\unc{0.002} & $0.196$\unc{0.001} & $0.114$\unc{0.002} & $0.228$\unc{0.005} & $0.240$\unc{0.003} & $0.190$\unc{0.004} & $0.105$\unc{0.002} & $0.216$\unc{0.008} & $0.234$\unc{0.002} \\
 & NTA+GR & $0.251$\unc{0.004} & $0.108$\unc{0.003} & $0.308$\unc{0.004} & $0.257$\unc{0.002} & $0.243$\unc{0.006} & $0.104$\unc{0.002} & $0.298$\unc{0.005} & $0.239$\unc{0.003} & $0.262$\unc{0.005} & $0.109$\unc{0.002} & $0.296$\unc{0.002} & $0.255$\unc{0.001} & $0.257$\unc{0.006} & $0.104$\unc{0.003} & $0.295$\unc{0.008} & $0.245$\unc{0.001} \\
 & NTA+CA & $0.259$\unc{0.002} & $0.106$\unc{0.002} & $0.316$\unc{0.008} & $0.254$\unc{0.002} & $0.254$\unc{0.004} & $0.102$\unc{0.002} & $0.307$\unc{0.005} & $0.239$\unc{0.004} & $0.271$\unc{0.007} & $0.107$\unc{0.002} & $0.302$\unc{0.007} & $0.253$\unc{0.001} & $0.266$\unc{0.010} & $0.103$\unc{0.001} & $0.304$\unc{0.003} & $0.242$\unc{0.001} \\
 & FedDAT & $0.235$\unc{0.001} & $0.113$\unc{0.002} & $0.274$\unc{0.005} & $0.254$\unc{0.004} & $0.214$\unc{0.010} & $0.116$\unc{0.002} & $0.266$\unc{0.010} & $0.238$\unc{0.004} & $0.229$\unc{0.009} & $0.114$\unc{0.001} & $0.281$\unc{0.006} & $0.252$\unc{0.001} & $0.226$\unc{0.007} & $0.109$\unc{0.002} & $0.263$\unc{0.012} & $0.259$\unc{0.004} \\
 & FedAdapt\_no\_CA & $0.213$\unc{0.005} & $0.111$\unc{0.002} & $0.246$\unc{0.004} & $0.254$\unc{0.002} & $0.189$\unc{0.003} & $0.114$\unc{0.001} & $0.237$\unc{0.008} & $0.235$\unc{0.004} & $0.206$\unc{0.009} & $0.111$\unc{0.001} & $0.255$\unc{0.006} & $0.252$\unc{0.001} & $0.201$\unc{0.011} & $0.108$\unc{0.002} & $0.234$\unc{0.008} & $0.262$\unc{0.004} \\
 & FedAdapt\_no\_CA+GR & $0.280$\unc{0.011} & $0.113$\unc{0.002} & $0.321$\unc{0.006} & $0.264$\unc{0.005} & $0.260$\unc{0.009} & $0.116$\unc{0.003} & $0.315$\unc{0.005} & $0.255$\unc{0.004} & $0.272$\unc{0.006} & $0.113$\unc{0.002} & $0.330$\unc{0.004} & $0.263$\unc{0.004} & $0.269$\unc{0.001} & $0.110$\unc{0.003} & $0.313$\unc{0.014} & $0.270$\unc{0.004} \\
 & FedAdapt & {\boldmath$0.289$}\uncbold{0.004} & {\boldmath$0.118$}\uncbold{0.003} & {\boldmath$0.329$}\uncbold{0.006} & {\boldmath$0.267$}\uncbold{0.003} & {\boldmath$0.271$}\uncbold{0.007} & {\boldmath$0.122$}\uncbold{0.003} & {\boldmath$0.323$}\uncbold{0.001} & {\boldmath$0.256$}\uncbold{0.003} & {\boldmath$0.282$}\uncbold{0.001} & {\boldmath$0.119$}\uncbold{0.001} & {\boldmath$0.337$}\uncbold{0.004} & {\boldmath$0.266$}\uncbold{0.003} & {\boldmath$0.279$}\uncbold{0.006} & {\boldmath$0.113$}\uncbold{0.003} & {\boldmath$0.321$}\uncbold{0.008} & {\boldmath$0.273$}\uncbold{0.002} \\
\hline
\Block{7-1}{\textbf{50\%}} & NTA & $0.185$\unc{0.006} & $0.108$\unc{0.002} & $0.226$\unc{0.005} & $0.235$\unc{0.003} & $0.163$\unc{0.007} & $0.106$\unc{0.002} & $0.212$\unc{0.010} & $0.205$\unc{0.004} & $0.189$\unc{0.001} & $0.110$\unc{0.002} & $0.221$\unc{0.009} & $0.232$\unc{0.003} & $0.183$\unc{0.002} & $0.101$\unc{0.002} & $0.207$\unc{0.006} & $0.227$\unc{0.002} \\
 & NTA+GR & $0.243$\unc{0.003} & $0.105$\unc{0.002} & $0.298$\unc{0.005} & $0.248$\unc{0.005} & $0.234$\unc{0.001} & $0.101$\unc{0.002} & $0.288$\unc{0.004} & $0.231$\unc{0.001} & $0.253$\unc{0.010} & $0.106$\unc{0.001} & $0.286$\unc{0.004} & $0.246$\unc{0.005} & $0.248$\unc{0.003} & $0.101$\unc{0.002} & $0.284$\unc{0.002} & $0.237$\unc{0.003} \\
 & NTA+CA & $0.251$\unc{0.006} & $0.103$\unc{0.002} & $0.305$\unc{0.009} & $0.246$\unc{0.002} & $0.245$\unc{0.004} & $0.099$\unc{0.003} & $0.296$\unc{0.010} & $0.231$\unc{0.002} & $0.262$\unc{0.010} & $0.103$\unc{0.002} & $0.293$\unc{0.003} & $0.244$\unc{0.001} & $0.257$\unc{0.009} & $0.099$\unc{0.002} & $0.293$\unc{0.003} & $0.234$\unc{0.004} \\
 & FedDAT & $0.228$\unc{0.007} & $0.109$\unc{0.002} & $0.265$\unc{0.002} & $0.245$\unc{0.004} & $0.207$\unc{0.003} & $0.112$\unc{0.002} & $0.256$\unc{0.004} & $0.229$\unc{0.002} & $0.222$\unc{0.005} & $0.110$\unc{0.002} & $0.272$\unc{0.002} & $0.243$\unc{0.001} & $0.218$\unc{0.007} & $0.105$\unc{0.001} & $0.254$\unc{0.003} & $0.250$\unc{0.004} \\
 & FedAdapt\_no\_CA & $0.206$\unc{0.008} & $0.107$\unc{0.002} & $0.237$\unc{0.009} & $0.246$\unc{0.005} & $0.182$\unc{0.002} & $0.110$\unc{0.002} & $0.228$\unc{0.005} & $0.226$\unc{0.003} & $0.199$\unc{0.011} & $0.107$\unc{0.001} & $0.246$\unc{0.008} & $0.243$\unc{0.004} & $0.194$\unc{0.001} & $0.105$\unc{0.001} & $0.225$\unc{0.005} & $0.253$\unc{0.002} \\
 & FedAdapt\_no\_CA+GR & $0.271$\unc{0.002} & $0.109$\unc{0.002} & $0.311$\unc{0.011} & $0.256$\unc{0.003} & $0.251$\unc{0.008} & $0.112$\unc{0.002} & $0.304$\unc{0.007} & $0.246$\unc{0.002} & $0.263$\unc{0.005} & $0.110$\unc{0.002} & $0.319$\unc{0.005} & $0.254$\unc{0.004} & $0.260$\unc{0.011} & $0.107$\unc{0.002} & $0.302$\unc{0.013} & $0.261$\unc{0.001} \\
 & FedAdapt & {\boldmath$0.280$}\uncbold{0.005} & {\boldmath$0.114$}\uncbold{0.003} & {\boldmath$0.318$}\uncbold{0.005} & {\boldmath$0.258$}\uncbold{0.004} & {\boldmath$0.261$}\uncbold{0.002} & {\boldmath$0.118$}\uncbold{0.002} & {\boldmath$0.312$}\uncbold{0.003} & {\boldmath$0.247$}\uncbold{0.003} & {\boldmath$0.272$}\uncbold{0.003} & {\boldmath$0.115$}\uncbold{0.002} & {\boldmath$0.326$}\uncbold{0.007} & {\boldmath$0.257$}\uncbold{0.004} & {\boldmath$0.270$}\uncbold{0.001} & {\boldmath$0.109$}\uncbold{0.001} & {\boldmath$0.310$}\uncbold{0.006} & {\boldmath$0.264$}\uncbold{0.002} \\
\hline
\Block{7-1}{\textbf{25\%}} & NTA & $0.171$\unc{0.006} & $0.100$\unc{0.002} & $0.209$\unc{0.006} & $0.217$\unc{0.001} & $0.148$\unc{0.001} & $0.098$\unc{0.002} & $0.194$\unc{0.005} & $0.186$\unc{0.005} & $0.174$\unc{0.009} & $0.102$\unc{0.002} & $0.204$\unc{0.007} & $0.213$\unc{0.005} & $0.168$\unc{0.001} & $0.092$\unc{0.002} & $0.189$\unc{0.006} & $0.210$\unc{0.002} \\
 & NTA+GR & $0.224$\unc{0.001} & $0.096$\unc{0.002} & $0.275$\unc{0.010} & $0.229$\unc{0.007} & $0.214$\unc{0.005} & $0.093$\unc{0.002} & $0.264$\unc{0.003} & $0.211$\unc{0.002} & $0.233$\unc{0.009} & $0.097$\unc{0.002} & $0.265$\unc{0.007} & $0.227$\unc{0.001} & $0.228$\unc{0.005} & $0.092$\unc{0.002} & $0.261$\unc{0.007} & $0.219$\unc{0.003} \\
 & NTA+CA & $0.232$\unc{0.004} & $0.095$\unc{0.002} & $0.282$\unc{0.005} & $0.227$\unc{0.005} & $0.224$\unc{0.001} & $0.091$\unc{0.003} & $0.272$\unc{0.008} & $0.211$\unc{0.001} & $0.241$\unc{0.004} & $0.095$\unc{0.001} & $0.270$\unc{0.002} & $0.225$\unc{0.006} & $0.237$\unc{0.004} & $0.091$\unc{0.002} & $0.269$\unc{0.011} & $0.217$\unc{0.001} \\
 & FedDAT & $0.210$\unc{0.010} & $0.101$\unc{0.002} & $0.244$\unc{0.003} & $0.226$\unc{0.005} & $0.189$\unc{0.007} & $0.104$\unc{0.002} & $0.236$\unc{0.008} & $0.210$\unc{0.004} & $0.205$\unc{0.001} & $0.102$\unc{0.002} & $0.251$\unc{0.003} & $0.225$\unc{0.003} & $0.201$\unc{0.011} & $0.097$\unc{0.003} & $0.233$\unc{0.009} & $0.231$\unc{0.005} \\
 & FedAdapt\_no\_CA & $0.190$\unc{0.003} & $0.099$\unc{0.001} & $0.219$\unc{0.005} & $0.227$\unc{0.001} & $0.166$\unc{0.002} & $0.102$\unc{0.001} & $0.209$\unc{0.007} & $0.206$\unc{0.005} & $0.183$\unc{0.010} & $0.099$\unc{0.002} & $0.228$\unc{0.006} & $0.224$\unc{0.003} & $0.179$\unc{0.004} & $0.097$\unc{0.002} & $0.206$\unc{0.009} & $0.234$\unc{0.002} \\
 & FedAdapt\_no\_CA+GR & $0.250$\unc{0.003} & $0.101$\unc{0.002} & $0.287$\unc{0.005} & $0.236$\unc{0.004} & $0.231$\unc{0.005} & $0.104$\unc{0.002} & $0.280$\unc{0.001} & $0.226$\unc{0.005} & $0.243$\unc{0.007} & $0.101$\unc{0.002} & $0.295$\unc{0.004} & $0.235$\unc{0.004} & $0.240$\unc{0.008} & $0.098$\unc{0.002} & $0.278$\unc{0.007} & $0.242$\unc{0.006} \\
 & FedAdapt & {\boldmath$0.258$}\uncbold{0.001} & {\boldmath$0.105$}\uncbold{0.001} & {\boldmath$0.293$}\uncbold{0.005} & {\boldmath$0.238$}\uncbold{0.005} & {\boldmath$0.240$}\uncbold{0.011} & {\boldmath$0.109$}\uncbold{0.002} & {\boldmath$0.287$}\uncbold{0.008} & {\boldmath$0.227$}\uncbold{0.005} & {\boldmath$0.251$}\uncbold{0.008} & {\boldmath$0.106$}\uncbold{0.002} & {\boldmath$0.302$}\uncbold{0.009} & {\boldmath$0.237$}\uncbold{0.002} & {\boldmath$0.249$}\uncbold{0.003} & {\boldmath$0.100$}\uncbold{0.002} & {\boldmath$0.285$}\uncbold{0.006} & {\boldmath$0.244$}\uncbold{0.003} \\
\hline
\Block{2-1}{\textbf{--}} & \stack{Centralized}{(Hypothetical Upper-Bound)} & $0.301$\unc{0.002} & $0.122$\unc{0.001} & $0.347$\unc{0.002} & $0.288$\unc{0.002} & $0.281$\unc{0.002} & $0.125$\unc{0.001} & $0.340$\unc{0.002} & $0.276$\unc{0.002} & $0.294$\unc{0.002} & $0.122$\unc{0.001} & $0.355$\unc{0.002} & $0.287$\unc{0.002} & $0.291$\unc{0.002} & $0.119$\unc{0.001} & $0.338$\unc{0.002} & $0.294$\unc{0.002} \\
 & Local-only & $0.105$\unc{0.001} & $0.063$\unc{0.002} & $0.132$\unc{0.003} & $0.137$\unc{0.002} & $0.096$\unc{0.001} & $0.061$\unc{0.002} & $0.124$\unc{0.003} & $0.120$\unc{0.002} & $0.110$\unc{0.001} & $0.064$\unc{0.002} & $0.126$\unc{0.003} & $0.135$\unc{0.002} & $0.107$\unc{0.001} & $0.059$\unc{0.002} & $0.121$\unc{0.003} & $0.129$\unc{0.002} \\
\hline
\multicolumn{18}{c}{} \\
\hline
\Block{1-18}{\textbf{(b) Modality Feature Alignment:} $\mathrm{CKA}(\pi^{\mathrm{Vis}}, \pi^{\mathrm{Lid}})$ \textbf{Relative to the Pre-Trained Model}} & & & & & & & & & & & & & & & & & \\
\midrule
\textbf{Setting} & \textbf{Method} & \Block{1-4}{\textbf{3D Seg. Arrives}} & & & & \Block{1-4}{\textbf{3D Det. Arrives}} & & & & \Block{1-4}{\textbf{2D Seg. Arrives}} & & & & \Block{1-4}{\textbf{2D Det. Arrives}} & &\\
\midrule
\Block{2-1}{\textbf{Fed.}} & Fine-tuned & \Block{1-4}{0.86} & & & & \Block{1-4}{0.75} & & & & \Block{1-4}{0.80} & & & & \Block{1-4}{0.76} & &\\
\cline{2-18}
 & Fine-tuned + Task Adapter & \Block{1-4}{\textbf{0.89}} & & & & \Block{1-4}{\textbf{0.78}} & & & & \Block{1-4}{\textbf{0.83}} & & & & \Block{1-4}{\textbf{0.81}} & &\\
\bottomrule
\end{NiceTabular}}}
\end{table*}

These findings highlight the potential of M3T FedFMs for vehicular network operators, as they suggest that operators can leverage auxiliary, more general tasks collected across heterogeneous regions to enhance learning efficiency and performance on critical, domain/vehicular-specific objectives, such as traffic understanding or incident reasoning, without requiring the centralization of raw data.

\section{Conclusion}
In this paper, we presented a forward-looking vision for the deployment of M3T FedFMs in vehicular networks. We first introduced a modular architectural perspective for M3T FedFMs along with their training/fine-tuning paradigms. We then explored representative use cases of M3T FedFMs in vehicular networks across three key stakeholder groups: autonomous vehicles, human drivers and passengers, and vehicular network operators. 
Furthermore, we outlined a set of open challenges and research directions, aiming to guide future work toward scalable, privacy-aware, and intelligent vehicular systems powered by M3T FedFMs. 
Finally, we presented a case study to demonstrate the potential of M3T FedFMs in vehicular environments and publicly released our implementation to catalyze research in this area. 

\section{Acknowledgments}
This work was supported in part
by the U.S. National Science Foundation (NSF) under Grants ECCS-2543754, SaTC-2513164,
and ECCS-2512911.
\bibliographystyle{ieeetr}
\bibliography{References.bib}

\vspace{-12mm}
\begin{IEEEbiographynophoto}{Kasra Borazjani} is a Ph.D. student in the Department of Electrical Engineering at the University at Buffalo--SUNY, USA.
\end{IEEEbiographynophoto}
\vspace{-9.5mm}
\begin{IEEEbiographynophoto}{Fardis Nadimi} is a Ph.D. student in the Department of Electrical Engineering at the University at Buffalo--SUNY, USA.
\end{IEEEbiographynophoto}
\vspace{-9.5mm}
\begin{IEEEbiographynophoto}{Payam Abdisarabshali} is a Ph.D. student in the Department of Electrical Engineering at the University at Buffalo--SUNY, USA.
\end{IEEEbiographynophoto}
\vspace{-9.5mm}
\begin{IEEEbiographynophoto}{Owen Palinski} is an M.Sc. student in the Department of Electrical Engineering at the University at Buffalo--SUNY, USA.
\end{IEEEbiographynophoto}
\vspace{-9.5mm}
\begin{IEEEbiographynophoto}{Allan Salihovic} is a Ph.D. student in the Department of Electrical Engineering at the University at Buffalo--SUNY, USA.
\end{IEEEbiographynophoto}
\vspace{-9.5mm}
\begin{IEEEbiographynophoto}{Dinh Nguyen} is an  assistant professor of Electrical and Computer Engineering at the University of Alabama--Huntsville, USA.
\end{IEEEbiographynophoto}
\vspace{-9.5mm}
\begin{IEEEbiographynophoto}{Minghui Liwang} is an associate professor of Electrical and Computer Engineering at Tongji University, China.
\end{IEEEbiographynophoto}
\vspace{-9.5mm}
\begin{IEEEbiographynophoto}{Seyyedali Hosseinalipour} is an assistant professor of Electrical Engineering at the University at Buffalo--SUNY, USA.
\end{IEEEbiographynophoto}

\end{document}